# Evolution of Visual Odometry Techniques

Shashi Poddar, Rahul Kottath, Vinod Karar

*Abstract*— With rapid advancements in the area of mobile robotics and industrial automation, a growing need has arisen towards accurate navigation and localization of moving objects. Camera based motion estimation is one such technique which is gaining huge popularity owing to its simplicity and use of limited resources in generating motion path. In this paper, an attempt is made to introduce this topic for beginners covering different aspects of vision based motion estimation task. The evolution of VO schemes over last few decades is discussed under two broad categories, that is, geometric and non-geometric approaches. The geometric approaches are further detailed under three different classes, that is, feature-based, appearance-based, and a hybrid of feature and appearance based schemes. The non-geometric approach is one of the recent paradigm shift from conventional pose estimation technique and is thus discussed in a separate section. Towards the end, a list of different datasets for visual odometry and allied research areas are provided for a ready reference.

## I. INTRODUCTION

With rising automation in different engineering fields, mobile robotics is gaining huge popularity. The unmanned vehicle is one such proliferating example that is expanding its fleet in different applications ranging from commercial to strategic use. One of the simplest mechanism to estimate the motion of a terrestrial vehicle is to use wheel encoders. However, these have limited usage in ground vehicles and suffer from inaccuracies that occur due to wheel slippage during movement in muddy, slippery, sandy or loose terrains. The errors arising at each instant gets accumulated over time and the estimated pose drifts in proportion to the distance traveled [1]. Traditional navigation approaches such as inertial navigation systems (INS), the global positioning system (GPS), SONAR, RADAR, and LIDAR are currently in use for different applications [2]. Unavailability of GPS signals in an indoor and under-surface environment, unacceptable high drift using inertial sensors during extended GPS outages, issues of possible confusion with nearby robots for SONAR & RADAR, and the line of sight requirement for laser-based systems are some of the limitations associated with these navigation systems. One of the promising solutions lies in the art of visual odometry that helps in estimating motion information with the help of cameras mounted over the vehicle.

The onboard vision system tracks visual landmarks to estimate rotation and translation between two-time instants. The art of vision-based navigation is inspired by the behavior of a bird which relies heavily on its vision for guidance and control [3]. The initial works on estimating motion from a camera by Moravec has helped in establishing the current visual odometry (VO) pipeline [4]. Simultaneous localization and mapping (SLAM), a superset of VO, localizes and builds a map of its environment along with the trajectory of a moving object [5]. However, our discussion in this paper is limited to visual odometry, which incrementally estimates the camera pose and refines it using optimization technique. A visual odometry system consists of a specific camera arrangement, the software architecture and the hardware platform to yield camera pose at every time instant. The camera pose estimation can be either appearance or feature based. The appearance-based techniques operate on intensity values directly and matches template of sub-images over two frame or the optical flow values to estimate motion [6]. The feature-based techniques extract distinct interest points that can be tracked with the help of vectors describing the local region around the key-points. These techniques are dependent on the image texture and are generally not applicable in texture-less or low texture environments such as sandy soil, asphalt, etc. [7]. The VO technique can also be classified as geometric and learning based. The geometric VO techniques are the ones that explore camera geometry for estimating motion whereas the learning based VO scheme trains regression model to estimate motion parameter when fed with labeled data [8]. The learning-based VO technique does not require the camera parameters to be known initially and can estimate trajectories with correct scale even for monocular cases [9].

The VO scheme can be implemented either with a monocular, stereo, or RGB-D camera depending on the system design. Stereo VO mimics the human vision system and can estimate the image scale immediately unlike monocular VO. However, stereo camera systems require more calibration effort and stringent camera synchronization without which the error propagates over time. The monocular camera is preferred for inexpensive and small form factor applications such as phone, laptop, etc. where the mounting of two cameras with a specified baseline is not always feasible. Some of the approaches that aimed to recover scale information for monocular VO are the usage of IMU information [10], optimization approach during loop closure [11], and incorporating known dimensional information from walls, buildings, etc. [12]. An RGB-D camera provides color and depth information for each pixel in an image. The RGB-D VO starts with the 3D position of feature points which are then used to obtain transformation through iterative closest point algorithm [13]. The VO scheme has found its major application in the automobile industry in driver assistance and autonomous navigation [14]. One of the applications of visual

Shashi Poddar is with CSIR - Central Scientific Instruments Organisation, Chandigarh, India (email: shashipoddar@csio.res.in)

Rahul Kottath is with Academy of Scientific & Innovative Research, CSIR - CSIO Campus, Chandigarh India, (email: rahulkottath@gmail.com)

Vinod Karar is with CSIR - Central Scientific Instruments Organisation, Chandigarh, India (email: vinodkarar@csio.res.in)

odometry has been to estimate vehicle motion from the rear-parking camera and use this information with GPS to provide accurate localization [15]. The task of visual servoing (Moving the camera to a desired orientation) is very similar to the visual odometry problem requiring pose estimation for a different purpose [16]. These schemes are not only useful for navigation of rovers on surfaces of other planets such as Mars [17] but are also useful for tracking of satellites that needs to be repaired using a servicer [18].

Although these VO techniques have shown promising results for variety of these applications, they are sensitive to environmental changes such as lighting conditions, surrounding texture, the presence of water, snow, etc. Some of the other conditions that lead to poor tracking data are motion blur, the presence of shadows, visual similarity, degenerate configuration, and occlusions. Along with these, some man-made errors also creep into the data during image acquisition and processing steps such as lens distortion and calibration, feature matching, triangulation, trajectory drift due to dead-reckoning which lead to outliers. Therefore, the VO schemes need to be robust and have the ability to manage these issues efficiently. In order to handle the environmental conditions, different techniques have been proposed in the literature such as the usage of NIR cameras for dark environment [19] or usage of rank transform to handle lighting condition [20]. Kaess et al. handle data degeneration by dividing the image into two clusters based on disparity and computing rotation and translation with distant and nearby objects, respectively [21]. Several outlier rejection schemes have been proposed in the literature of which RANSAC and its different variants are very commonly used [22].

The drift in trajectory over image frames is compensated using different strategies such as loop closure and Bundle Adjustment (BA). SLAM is an extended Kalman filter (EKF) estimator that aims at obtaining accurate motion vector given all the past feature positions and their tracking information [23]. Unlike SLAM that reduces drift by loop closure detection while visiting same scene locations, bundle adjustment (BA) optimizes camera poses over image frames [24]. The BA framework minimizes the re-projection error over the observed 3D image point and the predicted image point obtained using camera pose, intrinsic camera parameter and distortion parameter [25]. Sliding BA uses a fixed window of previous image frames for optimization and is more popularly used for real-time applications. Alternately, the fusion of visual odometry with other motion estimation modalities such as IMU, GPS [26], absolute sensor, compass [6] also exist in literature and are used to improve position accuracy. The rest of this paper is divided into different sections. Section 2 details the evolution of visual odometry scheme under different sub-categories, that is, feature-based, appearance-based and learning-based along with some discussion on RGB-D based VO scheme. Section 3 provides a list of different datasets specific to visual odometry and their allied areas and finally section 4 concludes the paper.

## II. EVOLUTION OF VISUAL ODOMETRY

Visual odometry (VO) is defined as those set of algorithms that help in estimating motion by taking cues from the images. These sensors can be either monocular, stereo or RGB-D in nature and have different algorithm frameworks, respectively. VO has a wide range of applications varying from gaming & virtual reality, wearable computing, industrial manufacturing, healthcare, underwater, aerial, space robotics, driver assistance system, agriculture field robots, automobile, pedestrian & indoor navigation, and control & guidance of unmanned vehicle. In recent years, several VO techniques have been published in the literature and is a non-trivial task to have holistic view over the full breadth of these schemes. However, few judicious attempts have been made by some of the researchers in reviewing specific aspects of these approaches. One of the popular reviews in the area of motion from image sequences was presented by Aggarwal and Nandhakumar in 1988 by classifying them into feature-based and optical flow based [27]. Later, Sedouza and Kak surveyed the work carried out in last two decades and classified these techniques into map-based, map-building-based, and map-less navigation schemes [28]. The map-based navigation approach requires the robot to be fed with a model of the environment and a sequence of expected landmarks whereas the map-building scheme creates a representation of outer environment seen by the camera. Unlike these, the mapless approach does not require any map to be created for navigation and estimates motion by observing external objects [28]. Scaramuzza and Fraundorfer published two landmarks articles on feature based visual odometry pipeline which is very helpful to a newbie in this research area. It segregates the feature based pose estimation frameworks, into 2D-to-2D, 3D-to-3D, and 3D-to-2D in concise steps, providing details of their origin and implementation. In 2011, Weiss et al. classified the VO techniques based on the camera location, one in which it is placed in the environment and the other in which it is placed on the UAV, respectively [29]. The former techniques are good for accurate and robust motion estimation of a robot moving only in a known environment while the later tracks a known pattern or unknown landmark in the environment [30]. Aqel et al. segregated several topics of interest to VO research community such as use of different sensors, VO applications, approaches and their current limitation [1]. Yousif et al. attempted to provide an overview on structure from motion schemes which included visual odometry and other localization and mapping methods [31]. Recently, Janai et al. has put up a detailed review on different computer vision methodologies used for autonomous driving. It dedicates a sub-section to ego-motion estimation with brief overview of recent articles published in the area of stereo and monocular visual odometry [32]. Several articles have been reported in the literature that uses hybrid of two different approaches, sensing modality etc. For example, Scaramuzza & Siegwart proposes a hybrid of appearance and feature based approach [7], a combination of visual camera and LIDAR sensor [33], vision and IMU [26], vision and compass [6], etc. The information provided in this paper is derived from the above mentioned review works along with the other allied literature in the area of visual odometry. However, the discussion here is limited to the evolution of VO approach from its original form to its current scenario. In order to provide brevity, the evolution of VO has been covered under two broad sub-sections, that are, geometric and non-geometric-based approaches. These geometric approaches are the ones that exploit information from the projective geometry and the non-geometric approaches as the ones that are based on learning. The non-analytical approaches have gained recent popularity with the evolution of machine learning. The geometric approaches have been further classified as feature-based, appearance-based and a hybrid of both the feature and appearance based.

*A. Geometric Approaches: Feature-based*

The research on visual odometry found it's origin back in 1960's where a lunar rover was built by Stanford University for controlling it from the Earth. This cart was further explored by Moravec to demonstrate correspondence based stereo navigation approach in 1980 [34]. This scheme matches distinct feature between stereo images and triangulates them to the 3-D world frame. Once the robot moves, these feature points are matched in the next frame to obtain corresponding 3D point and generate motion parameters. Mattheis and Shafer improved upon this technique by modeling the triangulation error as 3-D Gaussian distribution rather than scalar weights [35]. Arun et al. proposed a least square based approach for determining a transformation between 3-D point clouds [36]. Weng et al. proposed a matrix-weighted least square solution which performed remarkably better than unweighted or scalar-weighted solution [37]. It also proposes an iterative optimal scheme to obtain motion parameters and 3-D points. Some of the researchers employed Kalman filtering to estimate motion parameters by modeling the noise in image data by Gaussian distribution [38] [39]. Olson et al. solves the motion estimation problem through a maximum-likelihood formulation and mentions the use of an absolute orientation sensor to reduce the error growth rate for long distance navigation [40].

Different formulations to solve feature-based VO had been discussed in the literature before 2000 and is also present in the literature in review articles. Hence, not much of it will be discussed here and the main emphasis will be to present the improvements in feature-based VO schemes chronologically post 2000 era. In 2002, Se et al. proposed to use scale invariant feature, SIFT for tracking interest points across image frames and estimation ego-motion [41]. Improper image calibration, feature mismatch, image noise, and triangulation error are some of the contributing factor toward outliers during pose estimation. Several outlier rejection schemes for robust estimation such as RANSAC, MLESAC and their variants have been proposed in the literature. Some of the researchers aimed at estimating motion parameters without having a priori knowledge of camera calibration parameters [42]. However, as mentioned by Nister, knowing the intrinsic parameters in advance helps in obtaining more accurate and robust motion estimate especially for planar or near planar scenes [43]. Nister also provided the step-by-step motion estimation framework for both the monocular & stereo cases and coined the popular term 'visual odometry' in this paper [44]. Later, Engels et al. estimated pose using this five-point algorithm followed by a bundle adjustment based refinement strategy [45]. Se et al. applied Hough transform and RANSAC approach for obtaining a consistent set of matches over which the least square minimization is applied to obtain pose estimation. It has also contributed towards the use of smaller feature descriptor of size 16 with sufficient discriminative power to match features, motivating researchers for more efficient feature descriptors [46]. Tardif et al. estimated rotation using epipolar geometry and translation using 3D map while optimizing the current location alone, rather than all previous locations as done in bundle adjustment [47]. Extensive work has also been reported towards simultaneous localization and mapping (a superset of pose estimation) during this time and later, but is not discussed in this paper [48, 5].

Until this time, extensive work had been reported on different motion estimation sub-routines such as estimating motion from different camera types [49, 50], estimating calibration & essential matrix [51], and pose refinement using bundle adjustment [24], [52] which contributed towards accuracy improvement. With this growing confidence on feature-based VO techniques and its demonstration on ground vehicle navigation, it was used in navigating Mars exploration rover [17]. It then received renewed interest among the researchers and several improvements were proposed in the literature. Kaess et al. used flow information for segregating distant and close points and estimated rotation and translation from them separately [21]. Kalantari et al. proposed a pose estimation algorithm using three points and the knowledge of vertical direction obtained from IMU or the vanishing point [53]. However, this scheme was unable to provide closed-form solution and had singularity issues. Naroditsky et al. presented a closed form solution using similar three-plus-one algorithm by using vanishing point or gravitational vector as the reference [54]. Later, in 2011 Scaramuzza et al. proposed the 1-point algorithm for motion estimation by utilizing the non-holonomic constraints of wheeled vehicles which switches to the standard 5-point algorithm on detection of lesser inliers [55]. Lee et al. extended this work for multi-camera set-up by modelling it as a generalized camera and proposed a 2-point algorithm for obtaining the scale metric [56]. Song et al. proposed a multi-thread monocular visual odometry technique that does not require any assumption regarding the environment for scale estimation. Epipolar search in all these threads with insertion of persistent 3D points at key-frames helped in improving its accuracy and speed [57]. Persson et al. extended this approach by generalizing it for the stereo case. The 3D correspondences and the pose estimated by motion model is used to predict track position and in refinement process [58].

Badino et al. improves the positional accuracy of feature points by averaging its position over all its previous occurrences and use these integrated features for improving ego-motion accuracy [4]. Kreso & Segvic (2015) paid significance to the camera calibration parameters and corrects them by matching feature points from one frame to the other with available ground truth motion [59]. Cvisic & Petrovic used a combination of stereo and monocular VO for estimating rotation using five-point algorithm and translation by minimizing re-projection error as done for the stereo case. Rotation estimating through monocular case help in overcoming errors arising due to imperfect calibration while translation estimation through stereo case increases the accuracy [60]. Bellavia et al. proposed a key-frame selection strategy based on the existence of image points with sufficient displacement [61]. Liu et al. proposed an improvement over the RANSAC scheme by generating the hypothesis preferentially and using three best hypotheses to estimate motion [62]. Two different sub-categories of these feature-based VO, that is, usage of different feature descriptors and selection of a feature subset is covered below.

*1) Use of Different Features*

Feature-based visual odometry involves keypoint detection, description, and a matching process to establish corresponding image points which are then used for motion estimation. The traditional edge and corner feature detection strategies such as Moravec and Harris corner detectors were very popular initially and provided fast image correspondences. With the evolution of scale and transformation invariant feature extraction methodologies such as SIFT, SURF, ORB, BRISK, etc. these are more widely used as compared to simple corner detectors. Different visual odometry research articles have used different feature detection-description techniques, that are, Harris corner detector in [63], SIFT in [47], SURF in [56], Censure in [64], BRIEF in [58], SURF – SYBA in [65], and ORB in [66]. Given the wide gamut of feature detection techniques, it is non-trivial to select a technique that suits one's speed and accuracy requirement. Some of the research works evaluated different feature detector and descriptors for visual odometry task and act as a beacon for judicious feature selection based on available hardware resources and design criteria. Schmidt et al. compared the performance of different detector-descriptor pairs and highlighted the speedup achieved by a pairing of single scale feature detector with a reduced version of SURF descriptor [67]. Jiang et al. further extended this work by experimenting on a larger set of detector - descriptor pairs and datasets [68]. It is shown that BRISK detector-descriptor is robust against image changes and takes lesser time as compared to SIFT and SURF for visual odometry pipeline. Additionally, it proposes the use of multiple scale detectors only for extreme motions while using single scale detectors such as corner-based features to expedite processing while maintaining similar accuracy. Further, Chien et al. compared the performance of SIFT, SURF, ORB and A-KAZE features for the task of visual odometry and found the SURF based technique to yield maximum accuracy while the ORB features to be computationally cheaper at the cost of lower accuracy [69].

Although most of the feature-based VO techniques use point-features, very less work has been done with line features owing to its computational complexity. In 2013, Witt and Weltin proposed an iterative closest multiple lines algorithm to use line feature for pose estimation [70]. However, this scheme could not be applied for images with high texture and needed complement with the point-based features. Ojeda and Jimenez combined the point and line features in a probabilistic manner [66] rather than combining them directly as attempted by Koletschka et al. [71]. The probabilistic combination lead to an efficient solution with reduced effect of noisy measurements and an easy integration in the probabilistic mobile robotics [66].

*2) Features Selection*

Not only the selection of appropriate feature detection technique, researchers have devised mechanism by which only a portion of the detected features are used for improved pose estimation. Kitt et al. incorporateded a bucketing approach for feature selection wherein the image is divided into grids such that each grid contributes only a specified number of matches for further processing [63]. This approach reduces computational complexity and improves ego-motion accuracy with uniform feature distribution [72]. Cvisic & Petrovic classified features from each bucket into four different classes and selected strongest features from each class for motion estimation [60]. Maeztu et al. carries out the complete feature detection, description, and matching in corresponding grids obtained by bucketing. It not only helps in improving estimated motion by reducing outliers but act as a framework for parallel implementation in multi-core architectures [73]. Kitt et al. extended this technique by classifying features into moving and non-moving feature with the help of randomized decision tree followed with the bucketing technique to select features for motion estimation [74]. Zhou et al. used random fern classifier to segregate matchable from non-matchable points and computed essential matrix only from the matchable ones \cite [75]. The main disadvantage of these classifier based techniques is that they require training in advance and thus an online learning approach is needed to adapt in different situations [74].

Escalera et al. proposed a stereo VO technique that uses static features belonging to the ground surface only, thus reducing the total number of features being used for pose estimation [76]. Recently, Kottath et al. proposed an inertia constrained VO approach which selects only those features that follow the predicted motion model [77]. It is a simplified implementation of the technique proposed by Kostavelis on non-iterative outlier removal for stereo VO [78] and Wu et al. wherein smoothness motion constraint is used to reject outliers [79].

The pose estimated from these feature-based VO scheme is generally passed through a filtering or an optimization framework for improved motion estimate. Feature-based EKF-SLAM has been a popular technique for localization and mapping in the computer vision community which uses sparse interest points [48]. A dedicated set of researchers have also employed Kalman filter for motion estimation, accounting for noise robustness. Webb et al. employed tracked image feature points along with their epipolar constraints as measurement model for estimated states in the EKF framework. Not much has been further detailed on these filtering or optimization based techniques and the discussion will remain limited to pose estimation work alone. The following sub-section gives a brief of appearance-based VO scheme that has evolved in parallel to the feature-based ego-motion estimation schemes.

*B. Appearance based visual odometry*

Appearance-based visual odometry is another class of geometric approaches that does not rely on sparse features and estimates motion by optimizing the photometric error. Generally, the feature-based techniques are said to be noisy and the features need not necessarily be distinguishable from their surroundings in smoothly varying landscapes such as foggy environment or a sandy area [80]. The feature-matching step at times leads to wrong association which needs to be removed and can even be expensive if implemented using neural networks. Instead, these appearance-based techniques (also referred as direct methods) utilize information from the complete image, leading to robust ego-motion estimate even in low-textured environments [81]. Using whole images rather than few landmarks reduces aliasing issues associated with similar looking places, works even with smooth varying landscapes and are fast to implement [80]. The appearance-based techniques are generally of two types: region-based matching and optical flow based.

The region-based matching either can be achieved through correlation (template matching) or with the help of global appearance based descriptors and image alignment approaches. Correlation based techniques (also referred as template matching) for aligning images had been a widely researched area in the past using global invariant image representations or similarity measures. These schemes had several limitations which were overcome with the use of locally invariant similarity measure and global constraints [82]. The image alignment technique proposed by Irani and Anandan is able to estimate parametric 2D motion model for images acquired by sensors from different modalities [82]. Mandelbaum et al. extended this scheme for estimating 3D ego-motion, which is iteratively refined over the multi-resolution framework. It estimates pose for a batch of images that is bootstrapped with a priori motion estimate for speeding up estimation process. The a priori information can be either obtained from the previous batch or an external sensor or through a Kalman filter prediction [83]. Vatani et al. proposed a simple and practical approach to ego-motion estimation using constrained correlation based approach. Some of the modifications carried out over simple correlation approach are correlation mask size based on image height, mask location as per vehicle motion and feeding of a smaller predictor area in which the mask is matched [84]. Yu et al. extended this work by using a rotated template that helps in estimating both the translation and rotation between two frames [85]. Vatani et al. selected an appropriate template image from multiple-templates and used linear forward prediction filter to select window location for faster and accurate matching process [86]. Frederic Labrosse proposed a visual compass technique based on template-matching to estimate pixel displacement in the image captured by an Omni-directional camera looking at the environment [80]. Scaramuzza incorporated the visual compass scheme to estimate rotation as this scheme is robust to systematic errors from camera calibration and error accumulation due to integration over time [87]. Gonzalez et al. incorporated Labrosse's visual compass technique for rotation estimation along with traditional template matching approach to estimate translation using two different cameras pointing at the environment and the ground, respectively [6]. Recently, Aqel et al. proposed an adaptive template-matching scheme with reduced mask size and changing template position based on vehicle acceleration [88]. Several recent works are reported towards robust template matching techniques for other applications and can be extended for visual odometry problem as well.

Some of the efforts were also made towards the usage of global image appearance for registring images that can then be used for estimating ego-motion. Goecke et al. make use of Fourier-Mellin transformation [89] while Menegatti et al. use phase information from image's Fourier signature for estimating vehicle motion from these global descriptors [90]. The use of image registration techniques for motion analysis finds its mention in the article published by Lucas and Kanade in the early 80s [91]. A set of techniques use pre-stored image sequences for comparison with the current image and yield an indoor navigation estimate [92], [93]. Zhou et al. used histograms to describe the appearance of pre-stored image frames as templates which are then compared with the histogram of current image for recognizing vehicle's current location [94]. However, these schemes are not able to detect the orientation accurately which was depicted in the experiments carried out by Pajda & Hlavac [95]. Jogan and Leonardis correlated images using a combination of zero phase representation (ZPR) and eigen-space of oriented images to yield rotation invariance but were sensitive to noise and occlusions [96].

Comport et al. use reference stereo image pairs to yield dense correspondences for estimating 6DOF pose. This scheme is based on the quadrifocal relationship between image intensities and is robust to occlusions, inter-frame displacements, and illumination changes [97]. Comport et al. later extended his own work by designing a scheme that minimizes the intensity error between the entire image while overcoming the inter-frame overlap issue associated with region-based approaches [98]. Lovegrove et al. proposed an image alignment approach to estimate vehicle motion by taking the advantage of texture present on planar road surfaces [15]. Some of the other region based matching schemes use motion parallax to compute 3D translation and parametric transformation between two frames [99], [100]. Irani et al. mentions the advantage of decomposing the camera motion into parametric motion and parallax displacements rather than into translation and rotation. This plane-plus-parallax scheme is said to be more robust, stable and simpler than the optical flow based approaches as it requires solving smaller sets of linear equations [101].

These region-based schemes require a specific interest area to be defined in the image that needs to be matched through sufficient overlap with the other image. Further, the image registration process requires an optimization technique to minimize an objective function, which is generally subjected to local minima and divergence issues [98]. An inappropriate choice of argument minimization criterion and existence of independently moving objects are some of the major concerns that can be avoided with optical flow based VO schemes [102]. Optical flow is one of the fundamental principles that define ego-motion of an observer seen in an image as per Gibson's ecological optics [103]. The use of optical flow for estimating motion information is inspired by the biological cues used by insects for navigation purposes [104]. Some early attempts toward estimating ego-motion from optical flow were taken up by Clocksin (1978) [105],

Ullman (1979) [106], and Prazdny (1980) [107]. However, most of the techniques consider the scene to contain a single object or restricted the motion to be translatory with an assumption of only planar surfaces in the environment. The basic formulation of optical flow proposed by Horn and Schunck [108] gets violated in the presence of motion discontinuities and varying illumination [109]. Gilad Adiv solved the motion discontinuity issue by computing motion for each of the connected partitions of flow vectors. These segments are later grouped together to formulate a motion hypothesis that is compatible with all the segments in a group [102]. Black and Anandan proposed to use statistical frameworks that helped in estimating the motion of the majority of the pixels while eliminating outliers [110]. Several works have been reported towards estimating illumination by modelling it with a multiplicative/ additive factor or in a recursive framework. Kim et al. addresses both the illumination and motion discontinuity issue by integrating Black and Anandan's approach for handling motion discontinuity [110] with that of Gennert and Negahdaripour's illumination variation model [111] specifically designed for motion estimation task [109]. These optical flow based motion estimation methods also referred to as direct method, use complete image information that can be applied to recover global 2D or 3D motion models [112]. Giachetti et al. proposed a correlation based dense optical flow technique for estimating ego-motion of a car moving in usual streets with the help of a TV camera mounted parallel to the ground [113]. The correlation based optical flow with large masks helps in overcoming instabilities associated with computing derivatives and temporal filtering helps in reducing the disturbances due to shock and vibration by rejecting horizontal component of optical flow. However, this scheme is not reliable in the presence of independently moving objects, and movement through hilly, dense vegetation, and cluttered area. K. J. Hanna described an iterative approach to estimate camera motion directly through brightness derivative while using ego-motion and brightness constraint for refinement [114]. Corke et al. mentioned the significance of using omnidirectional cameras for estimating motion as it can retain features for longer duration and incorporates more information [104]. Hyslop and Humbert used the wide-field motion information from optical flow for estimating 6-DOF motion parameters and provided reliable information for navigation in an unknown environment [115].

In 2005, Campbell et al. proposed an optical flow based ego-motion estimation technique wherein the rotation is estimated using features that are far from the camera and translation using near-by features [116]. Some of the research works aimed at estimating motion in a controlled environment that cannot be generalized for outdoor conditions. For example, the technique proposed by Srinivasan obtained ego-motion for the set-up requiring camera to track changes in ceiling light pattern having limited scope [117]. Grabe et al. (2012a) demonstrated an optical flow based closed loop controlled UAV operation using onboard hardware alone. This scheme aims at continuous motion recovery rather than estimating frame-to-frame motion alone [118]. Grabe et al. (2012b) further extended their work by employing features that belong to a dominant plane alone to obtain improved velocity estimates [119]. Tykkala and Comport presented a direct stereo based SLAM method wherein the motion is estimated by direct image alignment [120]. The LSD-SLAM estimates depth at pixels with large intensity gradients and also estimates rigid body motion by aligning images based on the depth map. However, this scheme uses cues from both the stereo and monocular set-up and handles brightness changes in the image frame to yield better estimates [121]. Recently, Engel et al. proposed a direct sparse odometry scheme, which optimizes the photometric error in a framework similar to sparse bundle adjustment. It avoids the use of geometric prior used in feature-based approaches and uses all image points to achieve robustness [81]. Several works have also been reported in the literature recently that estimates ego-motion using a combination of feature and optical flows [122].

Optical flow has not only been used for estimating motion but also to help UAVs navigate by providing cues related to the presence of an obstacle in the vehicle path [123]. However, optical flow based scheme have their own limitations such as matching in texture-less surfaces (concrete, sand, etc.) and computational complexity. The RGB-D cameras are one of the frameworks that have lower computational complexity as it provides depth values for image points directly through a depth sensor embedded in the color camera. One set of techniques formulates the VO task as energy minimization problem [124] while the other ones estimate trajectory by classical registration techniques [125]. The photometric error formulation of direct method is combined with the error in dense map obtained from RGB-D sensor to formulate a cost function which can be solved using numerical optimization algorithm [126]. The registration based scheme can achieve alignment based on shapes, features, or surface normal projection [127]. Dryanovski et al. proposed a scheme for aligning 3D point against a global model using iterative closest point algorithm [13]. Li and Lee proposed a fast visual odometry scheme by selecting few salient points on the source frame for ICP and integrating intensity values in the correspondence estimation for improved ICP [127]. A brief review of related works can be seen in the article published by Kerl et al. which estimates motion by registering the two RGB-D images directly on the basis of photometric error [128]. Whelan et al. proposes a robust RGB-D based visual odometry scheme which helps in colored volumetric reconstruction of different scenes and is one of the latest work in this area [129].

*C. Geometric Approaches: Hybrid of feature and appearance based*

The hybrid algorithms for visual odometry takes advantage of both the direct (feature-based) and indirect methods (appearance-based). Feature-based schemes provide reliable data at the cost of certain loss of available information while appearance-based methods provide dense reconstruction exploiting total available data but has errors associated with few areas. Oliensis and Werman proposes an algorithm that combines direct and indirect schemes in one framework with the main aim to incorporate all the available data and improving motion estimate [130]. Morency and Gupta proposed a hybrid registration scheme that incorporates feature tracking information and optical flow constraints in one framework [131]. Scaramuzza et al. used appearance-based approach to estimate rotation while translation is estimated by features extracted from the ground plane [132]. Forster et al. proposed a semi-direct VO technique that obtains feature correspondence from direct motion estimation which is then incorporated as matched points for feature-based pose estimation [133]. The direct motion estimation scheme here uses sparse model-based image alignment to obtain feature correspondence and is termed semi-direct owing to its fusion with feature based VO approach. Silva et al. proposes a dense ego-motion estimation technique complemented with the feature based VO to obtain translation scale factor accurately and later refined with the Kalman filter [134]. Silva et al. extends it further by employing probabilistic correspondence for fully dense probabilistic stereo ego-motion and has mentioned it to be robust against difficult image scenarios [135].

*D. Non-Geometric Approaches*

With the evolution of better computing resources, machine learning techniques are being used for several real-time applications. ALVINN was one the initial attempts towards the usage of machine learning techniques to improve the performance of NAVLAB, the Carnegie Mellon autonomous navigation test vehicle using a three layer neural network in 1989 [136]. It was further improved by speeding it by 5-times using "on the fly" training approach in which the system imitates the human driver under actual driving conditions [137]. Learning can be used in any stage of the odometry pipeline such as features learning for better pose estimation [138], estimating feature correspondence [139], homography estimation [140], etc.

Visual odometry based on machine learning is one of the emerging techniques for motion estimation, as it does not require the camera calibration parameters to be known explicitly. The labelled data is used to train a regression/ classification model that can estimate the ego-motion once an input image sequence is provided. These non-geometric learning-based approaches can estimate translation to the correct scale and are robust against similar kind of noises with which it is trained. This switching from geometry-based to learning-based approach is one of the recent paradigm shift in the area of visual navigation. One of the initial works towards learning-based VO by Roberts et al. aimed at learning the platform velocity and turn rate from optical flow [141]. Guizilini and Ramos eliminated the use of geometric model by learning the effect of camera motion on image structure and vehicle dynamics. It used a coupled Gaussian process for supervised learning of ego-motion from optical flow information [9], [142]. They further extended their work for estimating linear and angular velocities by using optical flow information from a single camera along with Multiple-output Gaussian process framework (MOGP) [143]. Konda and Memisevic made use of convolutional neural network (CNN) based synchrony auto encoder for joint estimation of depth and motion parameters from single / multiple cameras [144]. This work was further extended for visual odometry application by estimating local changes in velocity and direction through the CNN architecture [138]. Mohanty et al. used deep CNN to extract high level features for estimating transformation between two time instants [145]. Xu et al. used large scale crowd dataset to predict vehicle ego-motion from its previous state estimates and instantaneous camera observations [146]. An improved CNN, that is, recurrent CNN, is used for achieving end-to-end pose estimation by learning geometrical features in a sequential manner [147]. The CNN structure has also been used in estimating scale for monocular visual odometry with the help of street masks used for ground plane estimation [148]. Peretroukhin et al. incorporated a CNN variant, that is, Bayesian CNN to track sun direction and incorporated it into the VO pipeline for improved ego-motion estimate [149]. Recently, Zhan et al. proposed a learning scheme for VO framework which uses training data of single-view depth information along with two-view odometry data [150]. Different model estimation schemes such as support vector machine, Gaussian process, fuzzy logic, etc. are also reported in the literature and holds great potential to be extended in the future [151] [152]. With this brief analysis, it can be concluded that the machine learning based VO techniques hold huge potential for further improvements in estimating accurate motion estimate.

The above sub-sections provide a brief overview of the evolution of visual odometry schemes varying from geometric to non-geometric approaches. Along with these, VO schemes have also been developed for infrared cameras but this paper does not provide much detail on them [153], [154]. With progress in different subroutines of VO scheme, better and faster VO schemes are bound to evolve. Some of the recent works have shown the new directions in the motion estimation task such as pose estimation using event-based camera [155], direct sparse odometry [81], large scale direct SLAM [121], robust real time VO using dense RGB-D camera [129], etc. This evolution of VO schemes will still continue and this overview is a very small attempt to present different dimensions in which the VO scheme currently deals.

III. VISUAL ODOMETRY DATASETS

With growing research in robotics and computer vision algorithms, it became very important to generate benchmarking datasets with ground-truth values that help in comparing one algorithm over the other. In this attempt, several datasets have been put up by the researchers publicly for comparing ego-motion estimation and its allied techniques. Among the visual odometry stereo datasets, Malaga [156], [157] and New college [158] datasets were some of the earliest dataset for mobile robot localization. The Karlsruhe dataset [159] could not be very popular as it had some acquisition issues. The KITTI vision benchmark [160] suite is one of the most popular datasets in the computer vision research, especially for visual odometry task and is used by several researchers to compare their pose estimation scheme. Some of the datasets aimed to provide additional sensors data for better comparisons with ground truth and targeting a relatively larger research community that works on vision-aided navigation. The Wean Hall dataset [161], Kagaru Airborne Stereo Dataset [162], EuroC MAV dataset [163], Oxford robotcar dataset [164] are some of the datasets that provides stereo camera frames as along with the information from LIDAR, IMU, and GPS. Of these, Oxford robotcar dataset is suited to deep learning-based schemes that require huge datasets for training and estimating motion through images directly. The TUM - monocular visual odometry dataset [165], [166] and LSD-SLAM [167] is dedicated to the development of pose estimation and localization through monocular camera. One of the recent datasets aiming at high-speed robotics is provided in Zurich – event-camera dataset [155] for designing new class of pose estimation algorithms with very high frame rate event-based cameras. Ford campus vision dataset [168], ETH - Vision \& Laser Datasets from a Heterogeneous UAV Fleet [169], Zurich urban micro aerial vehicle dataset [170], and TUM Visual-Inertial Dataset [171] are some of the datasets designed specifically for SLAM, collaborative 3D reconstruction, appearance based localization, and visual odometry based applications, respectively.

RGB-D based motion estimation is one of the other research areas that is gaining importance and thus dedicated benchmark datasets have also been put up for the same. The TUM-RGB-D SLAM dataset [172] published in 2012 is one of the earliest attempt towards providing RGB-D data for evaluation of visual odometry and visual SLAM schemes. The MIT SATA center dataset [173] and ICL- NUIM RGB-D dataset [174] maps the indoor environment with a RGB-D camera and focusses mainly on floor planning and surface reconstruction. Very recently, the ETH – RGB-D Dataset [175] has also been published which uses laser scanner to generate ground truth information for structure from motion kind of application.

Some of the ego-motion estimation schemes has also been developed and tested over synthetic datasets of which the New Tsukuba dataset [176] is very famous and used by several researchers. These datasets are generated entirely on computer using different photo editing software. The Multi-FoV synthetic dataset [177] is one of the latest attempt towards synthetic dataset that simulates flying robot hovering in a room and vehicle moving in a city.

IV. CONCLUSION

In this work, an attempt is made to provide a holistic picture of the visual odometry technique encompassing different branches of this tree. The paper starts with an introduction to the motion estimation schemes and their wide applications in different engineering fields. The VO schemes have been discussed under two broad categories, that is, geometric and non-geometric approaches. The gamut of geometric approach is very wide and is thus sub-divided into three different sub-class, that is, feature-based, appearance-based, and hybrid schemes. Towards the end, a list of different VO datasets is provided for ready reference which has been segregated into different classes depending on the sensing modality. On the basis of recent research articles, it is seen that a huge impetus is given towards machine learning based VO, RGB-D based VO and other hybrid schemes that takes the advantage of both direct and indirect/ sparse and dense approaches in one coherent framework. It is also pertinent to mention here that this work is not an exhaustive survey of visual odometry research articles as it is a growing research area and a huge amount of related work has happened in the past.

ACKNOWLEDGMENT

This research has been supported by DRDO - Aeronautical Research & Development Board through grant-in-aid project on "Design and development of Visual Odometry System.

REFERENCES

[1] M. O. A. Aqel, M. H. Marhaban, M. I. Saripan and N. B. Ismail, "Review of visual odometry: types, approaches, challenges, and applications," *SpringerPlus,* vol. 5, p. 1897, 2016.

[2] R. Madison, G. Andrews, P. DeBitetto, S. Rasmussen and M. Bottkol, "Vision-aided navigation for small UAVs in GPS-challenged environments," in *AIAA Infotech@ Aerospace 2007 Conference and Exhibit*, 2007.

[3] S. M. Ettinger, "Design and implementation of autonomous vision-guided micro air vehicles," 2001.

[4] H. Badino, A. Yamamoto and T. Kanade, "Visual odometry by multi-frame feature integration," in *Computer Vision Workshops (ICCVW), 2013 IEEE International Conference on*, 2013.

[5] H. Durrant-Whyte and T. Bailey, "Simultaneous localization and mapping: part I," *IEEE robotics \& automation magazine,* vol. 13, pp. 99-110, 2006.

[6] R. Gonzalez, F. Rodriguez, J. L. Guzman, C. Pradalier and R. Siegwart, "Combined visual odometry and visual compass for off-road mobile robots localization," *Robotica,* vol. 30, pp. 865-878, 2012.

[7] D. Scaramuzza and R. Siegwart, "Appearance-guided monocular omnidirectional visual odometry for outdoor ground vehicles," *IEEE transactions on robotics,* vol. 24, pp. 1015-1026, 2008.

[8] T. A. Ciarfuglia, G. Costante, P. Valigi and E. Ricci, "Evaluation of non-geometric methods for visual odometry," *Robotics and Autonomous Systems,* vol. 62, pp. 1717-1730, 2014.

[9] V. Guizilini and F. Ramos, "Visual odometry learning for unmanned aerial vehicles," in *Robotics and Automation (ICRA), 2011 IEEE International Conference on*, 2011.

[10] G. N{\"u}tzi, S. Weiss, D. Scaramuzza and R. Siegwart, "Fusion of IMU and vision for absolute scale estimation in monocular SLAM," *Journal of intelligent \& robotic systems,* vol. 61, pp. 287-299, 2011.

[11] H. Strasdat, J. M. M. Montiel and A. J. Davison, "Scale drift-aware large scale monocular SLAM," *Robotics: Science and Systems VI,* vol. 2, 2010.

[12] S. Hilsenbeck, A. M{\"o}ller, R. Huitl, G. Schroth, M. Kranz and E. Steinbach, "Scale-preserving long-term visual odometry for indoor navigation," in *Indoor Positioning and Indoor Navigation (IPIN), 2012 International Conference on*, 2012.

[13] I. Dryanovski, R. G. Valenti and J. Xiao, "Fast visual odometry and mapping from RGB-D data," in *Robotics and Automation (ICRA), 2013 IEEE International Conference on*, 2013.

[14] M. Bertozzi, A. Broggi and A. Fascioli, "Vision-based intelligent vehicles: State of the art and perspectives," *Robotics and Autonomous systems,* vol. 32, pp. 1-16, 2000.

[15] S. Lovegrove, A. J. Davison and J. Ibanez-Guzm{\'a}n, "Accurate visual odometry from a rear parking camera," in *Intelligent Vehicles Symposium (IV), 2011 IEEE*, 2011.

[16] A. I. Comport, E. Marchand, M. Pressigout and F. Chaumette, "Real-time markerless tracking for augmented reality: the virtual visual servoing framework," *IEEE Transactions on visualization and computer graphics,* vol. 12, pp. 615-628, 2006.

[17] M. Maimone, Y. Cheng and L. Matthies, "Two years of visual odometry on the mars exploration rovers," *Journal of Field Robotics,* vol. 24, pp. 169-186, 2007.

[18] N. W. Oumer and G. Panin, "3D point tracking and pose estimation of a space object using stereo images," in *Pattern Recognition (ICPR), 2012 21st International Conference on*, 2012.

[19] J. Ruppelt and G. F. Trommer, "Stereo-camera visual odometry for outdoor areas and in dark indoor environments," *IEEE Aerospace and Electronic Systems Magazine,* vol. 31, pp. 4-12, 2016.

[20] C. Golban, S. Istvan and S. Nedevschi, "Stereo based visual odometry in difficult traffic scenes," in *Intelligent Vehicles Symposium (IV), 2012 IEEE*, 2012.

[21] M. Kaess, K. Ni and F. Dellaert, "Flow separation for fast and robust stereo odometry," in *Robotics and Automation, 2009. ICRA'09. IEEE International Conference on*, 2009.

[22] R. Raguram, O. Chum, M. Pollefeys, J. Matas and J.-M. Frahm, "USAC: a universal framework for random sample consensus," *IEEE transactions on pattern analysis and machine intelligence,* vol. 35, pp. 2022-2038, 2013.

[23] A. J. Davison, "Real-time simultaneous localisation and mapping with a single camera," in *null*, 2003.

[24] B. Triggs, P. F. McLauchlan, R. I. Hartley and A. W. Fitzgibbon, "Bundle adjustment—a modern synthesis," in *International workshop on vision algorithms*, 1999.

[25] M. I. A. Lourakis and A. A. Argyros, "SBA: A software package for generic sparse bundle adjustment," *ACM Transactions on Mathematical Software (TOMS),* vol. 36, p. 2, 2009.

[26] M. Agrawal and K. Konolige, "Real-time localization in outdoor environments using stereo vision and inexpensive gps," in *Pattern Recognition, 2006. ICPR 2006. 18th International Conference on*, 2006.

[27] J. K. Aggarwal and N. Nandhakumar, "On the computation of motion from sequences of images-a review," *Proceedings of the IEEE,* vol. 76, pp. 917-935, 1988.

[28] G. N. DeSouza and A. C. Kak, "Vision for mobile robot navigation: A survey," *IEEE transactions on pattern analysis and machine intelligence,* vol. 24, pp. 237-267, 2002.

[29] S. Weiss, D. Scaramuzza and R. Siegwart, "Monocular-SLAM--based navigation for autonomous micro helicopters in GPS-denied environments," *Journal of Field Robotics,* vol. 28, pp. 854-874, 2011.

[30] D. Eynard, P. Vasseur, C. Demonceaux and V. Frémont, "Real time UAV altitude, attitude and motion estimation from hybrid stereovision," *Autonomous Robots,* vol. 33, pp. 157-172, 2012.

[31] K. Yousif, A. Bab-Hadiashar and R. Hoseinnezhad, "An overview to visual odometry and visual slam: Applications to mobile robotics," *Intelligent Industrial Systems,* vol. 1, pp. 289-311, 2015.

[32] J. Janai, F. Güney, A. Behl and A. Geiger, "Computer vision for autonomous vehicles: Problems, datasets and state-of-the-art," *arXiv preprint arXiv:1704.05519,* 2017.

[33] J. Zhang and S. Singh, "Visual-lidar odometry and mapping: Low-drift, robust, and fast," in *Robotics and Automation (ICRA), 2015 IEEE International Conference on*, 2015.

[34] H. P. Moravec, "Obstacle avoidance and navigation in the real world by a seeing robot rover.," 1980.

[35] L. Matthies and S. T. E. V. E. N. A. Shafer, "Error modeling in stereo navigation," *IEEE Journal on Robotics and Automation,* vol. 3, pp. 239-248, 1987.

[36] K. S. Arun, T. S. Huang and S. D. Blostein, "Least-squares fitting of two 3-D point sets," *IEEE Transactions on pattern analysis and machine intelligence,* pp. 698-700, 1987.

[37] J. Weng, P. Cohen and N. Rebibo, "Motion and structure estimation from stereo image sequences," *ieee Transactions on Robotics and Automation,* vol. 8, pp. 362-382, 1992.

[38] T. J. Broida and R. Chellappa, "Estimation of object motion parameters from noisy images," *IEEE transactions on pattern analysis and machine intelligence,* pp. 90-99, 1986.

[39] J. Hallam, "Resolving observer motion by object tracking," in *Proceedings of the Eighth international joint conference on Artificial intelligence-Volume 2*, 1983.

[40] C. F. Olson, L. H. Matthies, M. Schoppers and M. W. Maimone, "Stereo ego-motion improvements for robust rover navigation," in *Robotics and Automation, 2001. Proceedings 2001 ICRA. IEEE International Conference on*, 2001.

[41] S. Se, D. Lowe and J. Little, "Mobile robot localization and mapping with uncertainty using scale-invariant visual landmarks," *The international Journal of robotics Research,* vol. 21, pp. 735-758, 2002.

[42] R. I. Hartley, "Estimation of relative camera positions for uncalibrated cameras," in *European conference on computer vision*, 1992.

[43] D. Nistér, "An efficient solution to the five-point relative pose problem," *IEEE transactions on pattern analysis and machine intelligence,* vol. 26, pp. 756-770, 2004.

[44] D. Nistér, O. Naroditsky and J. Bergen, "Visual odometry," in *Computer Vision and Pattern Recognition, 2004. CVPR 2004. Proceedings of the 2004 IEEE Computer Society Conference on*, 2004.

[45] C. Engels, H. Stewénius and D. Nistér, "Bundle adjustment rules," *Photogrammetric computer vision,* vol. 2, 2006.

[46] S. Se, D. G. Lowe and J. J. Little, "Vision-based global localization and mapping for mobile robots," *IEEE Transactions on robotics,* vol. 21, pp. 364-375, 2005.

[47] J.-P. Tardif, Y. Pavlidis and K. Daniilidis, "Monocular visual odometry in urban environments using an omnidirectional camera," in *Intelligent Robots and Systems, 2008. IROS 2008. IEEE/RSJ International Conference on*, 2008.

[48] A. J. Davison and D. W. Murray, "Simultaneous localization and map-building using active vision," *IEEE transactions on pattern analysis and machine intelligence,* vol. 24, pp. 865-880, 2002.

[49] R. Hartley and A. Zisserman, Multiple view geometry in computer vision, Cambridge university press, 2003.

[50] P. Chang and M. Hebert, "Omni-directional structure from motion," in *Omnidirectional Vision, 2000. Proceedings. IEEE Workshop on*, 2000.

[51] B. Mičušík and T. Pajdla, "Omnidirectional camera model and epipolar geometry estimation by ransac with bucketing?," in *Scandinavian Conference on Image Analysis*, 2003.

[52] M. Lhuillier, "Automatic structure and motion using a catadioptric camera," in *Proceedings of the 6th Workshop on Omnidirectional Vision, Camera Networks and Non-Classical Cameras*, 2005.

[53] M. Kalantari, A. Hashemi, F. Jung and J.-P. Guédon, "A new solution to the relative orientation problem using only 3 points and the vertical direction," *Journal of Mathematical Imaging and Vision,* vol. 39, pp. 259-268, 2011.

[54] O. Naroditsky, X. S. Zhou, J. Gallier, S. I. Roumeliotis and K. Daniilidis, "Two efficient solutions for visual odometry using directional correspondence," *IEEE transactions on pattern analysis and machine intelligence,* vol. 34, pp. 818-824, 2012.

[55] D. Scaramuzza, "1-point-ransac structure from motion for vehicle-mounted cameras by exploiting non-holonomic constraints," *International journal of computer vision,* vol. 95, pp. 74-85, 2011.

[56] G. H. Lee, F. Faundorfer and M. Pollefeys, "Motion estimation for self-driving cars with a generalized camera," in *Computer Vision and Pattern Recognition (CVPR), 2013 IEEE Conference on*, 2013.

[57] S. Song, M. Chandraker and C. C. Guest, "Parallel, real-time monocular visual odometry," in *Robotics and Automation (ICRA), 2013 IEEE International Conference on*, 2013.

[58] M. Persson, T. Piccini, M. Felsberg and R. Mester, "Robust stereo visual odometry from monocular techniques," in *Intelligent Vehicles Symposium (IV), 2015 IEEE*, 2015.

[59] I. Krešo and S. Šegvic, "Improving the egomotion estimation by correcting the calibration bias," in *10th International Conference on Computer Vision Theory and Applications*, 2015.

[60] I. Cvi{\v{s}}i{\'c} and I. Petrovi{\'c}, "Stereo odometry based on careful feature selection and tracking," in *Mobile Robots (ECMR), 2015 European Conference on*, 2015.

[61] F. Bellavia, M. Fanfani and C. Colombo, "Selective visual odometry for accurate AUV localization," *Autonomous Robots*, vol. 41, pp. 133-143, 2017.

[62] Y. Liu, Y. Gu, J. Li and X. Zhang, "Robust Stereo Visual Odometry Using Improved RANSAC-Based Methods for Mobile Robot Localization," *Sensors*, vol. 17, p. 2339, 2017.

[63] B. Kitt, A. Geiger and H. Lategahn, "Visual odometry based on stereo image sequences with ransac-based outlier rejection scheme," in *Intelligent Vehicles Symposium (IV), 2010 IEEE*, 2010.

[64] K. Konolige, M. Agrawal and J. Sola, "Large-scale visual odometry for rough terrain," in *Robotics research*, Springer, 2010, pp. 201-212.

[65] A. Desai and D.-J. Lee, "Visual odometry drift reduction using SYBA descriptor and feature transformation," *IEEE Transactions on Intelligent Transportation Systems*, vol. 17, pp. 1839-1851, 2016.

[66] R. Gomez-Ojeda and J. Gonzalez-Jimenez, "Robust stereo visual odometry through a probabilistic combination of points and line segments," in *Robotics and Automation (ICRA), 2016 IEEE International Conference on*, 2016.

[67] A. Schmidt, M. Kraft and A. Kasi{\'n}ski, "An evaluation of image feature detectors and descriptors for robot navigation," in *International Conference on Computer Vision and Graphics*, 2010.

[68] Y. Jiang, Y. Xu and Y. Liu, "Performance evaluation of feature detection and matching in stereo visual odometry," *Neurocomputing*, vol. 120, pp. 380-390, 2013.

[69] H.-J. Chien, C.-C. Chuang, C.-Y. Chen and R. Klette, "When to use what feature? SIFT, SURF, ORB, or A-KAZE features for monocular visual odometry," in *Image and Vision Computing New Zealand (IVCNZ), 2016 International Conference on*, 2016.

[70] J. Witt and U. Weltin, "Robust stereo visual odometry using iterative closest multiple lines," in *Intelligent Robots and Systems (IROS), 2013 IEEE/RSJ International Conference on*, 2013.

[71] T. Koletschka, L. Puig and K. Daniilidis, "MEVO: Multi-environment stereo visual odometry," in *Intelligent Robots and Systems (IROS 2014), 2014 IEEE/RSJ International Conference on*, 2014.

[72] Z. Zhang, R. Deriche, O. Faugeras and Q.-T. Luong, "A robust technique for matching two uncalibrated images through the recovery of the unknown epipolar geometry," *Artificial intelligence*, vol. 78, pp. 87-119, 1995.

[73] L. De-Maeztu, U. Elordi, M. Nieto, J. Barandiaran and O. Otaegui, "A temporally consistent grid-based visual odometry framework for multi-core architectures," *Journal of Real-Time Image Processing*, vol. 10, pp. 759-769, 2015.

[74] B. Kitt, F. Moosmann and C. Stiller, "Moving on to dynamic environments: Visual odometry using feature classification," in *Intelligent Robots and Systems (IROS), 2010 IEEE/RSJ International Conference on*, 2010.

[75] W. Zhou, H. Fu and X. An, "A Classification-Based Visual Odometry Approach," in *Intelligent Human-Machine Systems and Cybernetics (IHMSC), 2016 8th International Conference on*, 2016.

[76] A. Escalera, E. Izquierdo, D. Mart{\'\i}n, B. Musleh, F. Garc{\'\i}a and J. M. Armingol, "Stereo visual odometry in urban environments based on detecting ground features," *Robotics and Autonomous Systems*, vol. 80, pp. 1-10, 2016.

[77] R. Kottath, D. P. Yalamandala, S. Poddar, A. P. Bhondekar and V. Karar, "Inertia constrained visual odometry for navigational applications," in *Image Information Processing (ICIIP), 2017 Fourth International Conference on*, 2017.

[78] I. Kostavelis, E. Boukas, L. Nalpantidis and A. Gasteratos, "Stereo-based visual odometry for autonomous robot navigation," *International Journal of Advanced Robotic Systems*, vol. 13, p. 21, 2016.

[79] M. Wu, S.-K. Lam and T. Srikanthan, "A Framework for Fast and Robust Visual Odometry," *IEEE Transactions on Intelligent Transportation Systems*, vol. 18, pp. 3433-3448, 2017.

[80] F. Labrosse, "The visual compass: Performance and limitations of an appearance-based method," *Journal of Field Robotics*, vol. 23, pp. 913-941, 2006.

[81] J. Engel, V. Koltun and D. Cremers, "Direct sparse odometry," *IEEE transactions on pattern analysis and machine intelligence*, vol. 40, pp. 611-625, 2018.

[82] M. Irani and P. Anandan, "Robust multi-sensor image alignment," in *Computer Vision, 1998. Sixth International Conference on*, 1998.

[83] R. Mandelbaum, G. Salgian and H. Sawhney, "Correlation-based estimation of ego-motion and structure from motion and stereo," in *Computer Vision, 1999. The Proceedings of the Seventh IEEE International Conference on*, 1999.

[84] N. Nourani-Vatani, J. Roberts and M. V. Srinivasan, "Practical visual odometry for car-like vehicles," in *Robotics and Automation, 2009. ICRA'09. IEEE International Conference on*, 2009.

[85] Y. Yu, C. Pradalier and G. Zong, "Appearance-based monocular visual odometry for ground vehicles," in *Advanced Intelligent Mechatronics (AIM), 2011 IEEE/ASME International Conference on*, 2011.

[86] N. Nourani-Vatani and P. V. K. Borges, "Correlation-based visual odometry for ground vehicles," *Journal of Field Robotics*, vol. 28, pp. 742-768, 2011.

[87] D. Scaramuzza, "Omnidirectional Vision," 2007.

[88] M. O. A. Aqel, M. H. Marhaban, M. I. Saripan and N. B. Ismail, "Adaptive-search template matching technique based on vehicle acceleration for monocular visual odometry system," *IEEJ Transactions on Electrical and Electronic Engineering*, vol. 11, pp. 739-752, 2016.

[89] R. Goecke, A. Asthana, N. Pettersson and L. Petersson, "Visual vehicle egomotion estimation using the fourier-mellin transform," in *Intelligent Vehicles Symposium, 2007 IEEE*, 2007.

[90] E. Menegatti, T. Maeda and H. Ishiguro, "Image-based memory for robot navigation using properties of omnidirectional images," *Robotics and Autonomous Systems*, vol. 47, pp. 251-267, 2004.

[91] B. Lucas and T. Kanade, "B.(1981). An iterative image registration technique with an application to stereo vision," in *Proc. DARPA Image Understanding Workshop*, 1981.

[92] Y. Matsumoto, M. Inaba and H. Inoue, "Visual navigation using view-sequenced route representation," in *Robotics and Automation, 1996. Proceedings., 1996 IEEE International Conference on*, 1996.

[93] T. Ohno, A. Ohya and S. Yuta, "Autonomous navigation for mobile robots referring pre-recorded image sequence," in *Intelligent Robots and Systems' 96, IROS 96, Proceedings of the 1996 IEEE/RSJ International Conference on*, 1996.

[94] C. Zhou, Y. Wei and T. Tan, "Mobile robot self-localization based on global visual appearance features," in *Robotics and Automation, 2003. Proceedings. ICRA'03. IEEE International Conference on*, 2003.

[95] T. Pajdla and V. Hlav{\'a}{\v{c}}, "Zero phase representation of panoramic images for image based localization," in *International Conference on Computer Analysis of Images and Patterns*, 1999.

[96] M. Jogan and A. Leonardis, "Robust localization using eigenspace of spinning-images," in *Omnidirectional Vision, 2000. Proceedings. IEEE Workshop on*, 2000.

[97] A. I. Comport, E. Malis and P. Rives, "Accurate quadrifocal tracking for robust 3d visual odometry," in *Robotics and Automation, 2007 IEEE International Conference on*, 2007.

[98] A. I. Comport, E. Malis and P. Rives, "Real-time quadrifocal visual odometry," *The International Journal of Robotics Research*, vol. 29, pp. 245-266, 2010.

[99] M. Irani, B. Rousso and S. Peleg, "Computing occluding and transparent motions," *International Journal of Computer Vision*, vol. 12, pp. 5-16, 1994.

[100] R. Cipolla, Y. Okamoto and Y. Kuno, "Robust structure from motion using motion parallax," in *Computer Vision, 1993. Proceedings., Fourth International Conference on*, 1993.

[101] M. Irani, B. Rousso and S. Peleg, "Recovery of ego-motion using region alignment," *IEEE Transactions on Pattern Analysis and Machine Intelligence,* vol. 19, pp. 268-272, 1997.

[102] G. Adiv, "Determining three-dimensional motion and structure from optical flow generated by several moving objects," *IEEE transactions on pattern analysis and machine intelligence,* pp. 384-401, 1985.

[103] J. J. Gibson, "Visually controlled locomotion and visual orientation in animals," *British journal of psychology,* vol. 49, pp. 182-194, 1958.

[104] M. V. Srinivasan, M. Lehrer, W. H. Kirchner and S. W. Zhang, "Range perception through apparent image speed in freely flying honeybees," *Visual neuroscience,* vol. 6, pp. 519-535, 1991.

[105] W. F. Clocksin, "Determining the orientation of surfaces from optical flow," in *Proceedings of the 1978 AISB/GI Conference on Artificial Intelligence*, 1978.

[106] S. Ullman, "The interpretation of structure from motion," *Proc. R. Soc. Lond. B,* vol. 203, pp. 405-426, 1979.

[107] K. Prazdny, "Egomotion and relative depth map from optical flow," *Biological cybernetics,* vol. 36, pp. 87-102, 1980.

[108] B. K. P. Horn and B. G. Schunck, "Determining optical flow," *Artificial intelligence,* vol. 17, pp. 185-203, 1981.

[109] Y.-H. Kim, A. M. Mart{\"i}nez and A. C. Kak, "Robust motion estimation under varying illumination," *Image and Vision Computing,* vol. 23, pp. 365-375, 2005.

[110] M. J. Black and P. Anandan, "The robust estimation of multiple motions: Parametric and piecewise-smooth flow fields," *Computer vision and image understanding,* vol. 63, pp. 75-104, 1996.

[111] M. A. Gennert and S. Negahdaripour, "Relaxing the brightness constancy assumption in computing optical flow," 1987.

[112] M. Irani and P. Anandan, "About direct methods," in *International Workshop on Vision Algorithms*, 1999.

[113] A. Giachetti, M. Campani and V. Torre, "The use of optical flow for road navigation," *IEEE transactions on robotics and automation,* vol. 14, pp. 34-48, 1998.

[114] K. J. Hanna, "Direct multi-resolution estimation of ego-motion and structure from motion," in *Visual Motion, 1991., Proceedings of the IEEE Workshop on*, 1991.

[115] A. M. Hyslop and J. S. Humbert, "Autonomous navigation in three-dimensional urban environments using wide-field integration of optic flow," *Journal of guidance, control, and dynamics,* vol. 33, pp. 147-159, 2010.

[116] J. Campbell, R. Sukthankar, I. Nourbakhsh and A. Pahwa, "A robust visual odometry and precipice detection system using consumer-grade monocular vision," in *Robotics and Automation, 2005. ICRA 2005. Proceedings of the 2005 IEEE International Conference on*, 2005.

[117] M. V. Srinivasan, "An image-interpolation technique for the computation of optic flow and egomotion," *Biological Cybernetics,* vol. 71, pp. 401-415, 1994.

[118] V. Grabe, H. H. B{\"u}lthoff and P. R. Giordano, "On-board velocity estimation and closed-loop control of a quadrotor UAV based on optical flow," in *Robotics and Automation (ICRA), 2012 IEEE International Conference on*, 2012.

[119] V. Grabe, H. H. B{\"u}lthoff and P. R. Giordano, "Robust optical-flow based self-motion estimation for a quadrotor UAV," in *Intelligent Robots and Systems (IROS), 2012 IEEE/RSJ International Conference on*, 2012.

[120] T. Tykk{\"a}l{\"a} and A. I. Comport, "A dense structure model for image based stereo slam," in *Robotics and Automation (ICRA), 2011 IEEE International Conference on*, 2011.

[121] J. Engel, J. St{\"u}ckler and D. Cremers, "Large-scale direct SLAM with stereo cameras," in *Intelligent Robots and Systems (IROS), 2015 IEEE/RSJ International Conference on*, 2015.

[122] R. Ranftl, V. Vineet, Q. Chen and V. Koltun, "Dense monocular depth estimation in complex dynamic scenes," in *Proceedings of the IEEE Conference on Computer Vision and Pattern Recognition*, 2016.

[123] S. Temizer, "Optical flow based local navigation," 2001.

[124] C. Kerl, J. Sturm and D. Cremers, "Dense visual SLAM for RGB-D cameras," in *Intelligent Robots and Systems (IROS), 2013 IEEE/RSJ International Conference on*, 2013.

[125] I. Dryanovski, C. Jaramillo and J. Xiao, "Incremental registration of RGB-D images," in *Robotics and Automation (ICRA), 2012 IEEE International Conference on*, 2012.

[126] T. Tykk{\"a}l{\"a}, C. Audras and A. I. Comport, "Direct iterative closest point for real-time visual odometry," in *Computer Vision Workshops (ICCV Workshops), 2011 IEEE International Conference on*, 2011.

[127] S. Li and D. Lee, "Fast visual odometry using intensity-assisted iterative closest point," *IEEE Robotics and Automation Letters,* vol. 1, pp. 992-999, 2016.

[128] C. Kerl, J. Sturm and D. Cremers, "Robust odometry estimation for RGB-D cameras," in *Robotics and Automation (ICRA), 2013 IEEE International Conference on*, 2013.

[129] T. Whelan, H. Johannsson, M. Kaess, J. J. Leonard and J. McDonald, "Robust real-time visual odometry for dense RGB-D mapping," in *Robotics and Automation (ICRA), 2013 IEEE International Conference on*, 2013.

[130] J. Oliensis and M. Werman, "Structure from motion using points, lines, and intensities," in *Computer Vision and Pattern Recognition, 2000. Proceedings. IEEE Conference on*, 2000.

[131] L.-P. Morency and R. Gupta, "Robust real-time egomotion from stereo images," in *Image Processing, 2003. ICIP 2003. Proceedings. 2003 International Conference on*, 2003.

[132] D. Scaramuzza, F. Fraundorfer, M. Pollefeys and R. Siegwart, "Closing the loop in appearance-guided structure-from-motion for omnidirectional cameras," in *The 8th Workshop on Omnidirectional Vision, Camera Networks and Non-classical Cameras-OMNIVIS*, 2008.

[133] C. Forster, M. Pizzoli and D. Scaramuzza, "SVO: Fast semi-direct monocular visual odometry," in *Robotics and Automation (ICRA), 2014 IEEE International Conference on*, 2014.

[134] H. Silva, A. Bernardino and E. Silva, "Probabilistic egomotion for stereo visual odometry," *Journal of Intelligent \& Robotic Systems,* vol. 77, pp. 265-280, 2015.

[135] H. Silva, A. Bernardino and E. Silva, "A voting method for stereo egomotion estimation," *International Journal of Advanced Robotic Systems,* vol. 14, p. 1729881417710795, 2017.

[136] D. A. Pomerleau, "Alvinn: An autonomous land vehicle in a neural network," in *Advances in neural information processing systems*, 1989.

[137] D. A. Pomerleau, "Efficient training of artificial neural networks for autonomous navigation," *Neural Computation,* vol. 3, pp. 88-97, 1991.

[138] K. R. Konda and R. Memisevic, "Learning Visual Odometry with a Convolutional Network.," in *VISAPP (1)*, 2015.

[139] D. DeTone, T. Malisiewicz and A. Rabinovich, "Deep image homography estimation," *arXiv preprint arXiv:1606.03798,* 2016.

[140] R. Memisevic, "Learning to relate images," *IEEE transactions on pattern analysis and machine intelligence,* vol. 35, pp. 1829-1846, 2013.

[141] R. Roberts, H. Nguyen, N. Krishnamurthi and T. Balch, "Memory-based learning for visual odometry," in *Robotics and Automation, 2008. ICRA 2008. IEEE International Conference on*, 2008.

[142] V. Guizilini and F. Ramos, "Semi-parametric models for visual odometry," in *Robotics and Automation (ICRA), 2012 IEEE International Conference on*, 2012.

[143] V. Guizilini and F. Ramos, "Semi-parametric learning for visual odometry," *The International Journal of Robotics Research,* vol. 32, pp. 526-546, 2013.

[144] K. Konda and R. Memisevic, "Unsupervised learning of depth and motion," *arXiv preprint arXiv:1312.3429,* 2013.

[145] V. Mohanty, S. Agrawal, S. Datta, A. Ghosh, V. D. Sharma and D. Chakravarty, "DeepVO: a deep learning approach for monocular visual odometry," *arXiv preprint arXiv:1611.06069,* 2016.

[146] H. Xu, Y. Gao, F. Yu and T. Darrell, "End-to-end learning of driving models from large-scale video datasets," *arXiv preprint,* 2016.

[147] S. Wang, R. Clark, H. Wen and N. Trigoni, "Deepvo: Towards end-to-end visual odometry with deep recurrent convolutional neural networks," in *Robotics and Automation (ICRA), 2017 IEEE International Conference on*, 2017.

[148] N. Fanani, A. St{\"u}rck, M. Ochs, H. Bradler and R. Mester, "Predictive monocular odometry (PMO): What is possible without RANSAC and multiframe bundle adjustment?," *Image and Vision Computing,* vol. 68, pp. 3-13, 2017.

[149] V. Peretroukhin, L. Clement and J. Kelly, "Inferring sun direction to improve visual odometry: A deep learning approach," *The International Journal of Robotics Research,* p. 0278364917749732, 2018.

[150] H. Zhan, R. Garg, C. S. Weerasekera, K. Li, H. Agarwal and I. Reid, "Unsupervised Learning of Monocular Depth Estimation and Visual Odometry with Deep Feature Reconstruction," *arXiv preprint arXiv:1803.03893,* 2018.

[151] M. Bojarski, D. Del Testa, D. Dworakowski, B. Firner, B. Flepp, P. Goyal, L. D. Jackel, M. Monfort, U. Muller, J. Zhang and others, "End to end learning for self-driving cars," *arXiv preprint arXiv:1604.07316,* 2016.

[152] R. Mahajan, P. V. Shanmuganathan, V. Karar and S. Poddar, "Flexible Threshold Visual Odometry Algorithm Using Fuzzy Logics," in *Proceedings of 2nd International Conference on Computer Vision \& Image Processing*, 2018.

[153] T. Mouats, N. Aouf, L. Chermak and M. A. Richardson, "Thermal stereo odometry for UAVs," *IEEE Sensors Journal,* vol. 15, pp. 6335-6347, 2015.

[154] T. Mouats, N. Aouf, A. D. Sappa, C. Aguilera and R. Toledo, "Multispectral stereo odometry," *IEEE Transactions on Intelligent Transportation Systems,* vol. 16, pp. 1210-1224, 2015.

[155] E. Mueggler, H. Rebecq, G. Gallego, T. Delbruck and D. Scaramuzza, "The event-camera dataset and simulator: Event-based data for pose estimation, visual odometry, and SLAM," *The International Journal of Robotics Research,* vol. 36, pp. 142-149, 2017.

[156] J.-L. Blanco, F.-A. Moreno and J. Gonzalez, "A collection of outdoor robotic datasets with centimeter-accuracy ground truth," *Autonomous Robots,* vol. 27, p. 327, 2009.

[157] J.-L. Blanco-Claraco, F.-{. Moreno-Due{\~n}as and J. Gonz{\'a}lez-Jim{\'e}nez, "The Málaga urban dataset: High-rate stereo and LiDAR in a realistic urban scenario," *The International Journal of Robotics Research,* vol. 33, pp. 207-214, 2014.

[158] M. Smith, I. Baldwin, W. Churchill, R. Paul and P. Newman, "The new college vision and laser data set," *The International Journal of Robotics Research,* vol. 28, pp. 595-599, 2009.

[159] A. Geiger, J. Ziegler and C. Stiller, "Stereoscan: Dense 3d reconstruction in real-time," in *Intelligent Vehicles Symposium (IV), 2011 IEEE*, 2011.

[160] A. Geiger, P. Lenz, C. Stiller and R. Urtasun, "Vision meets robotics: The KITTI dataset," *The International Journal of Robotics Research,* vol. 32, pp. 1231-1237, 2013.

[161] H. Alismail, B. Browning, M. B. Dias, B. Argall, B. Browning, Y. Gu, M. Veloso, B. Argall, B. Browning, M. Veloso and others, "Evaluating Pose Estimation Methods for Stereo Visual Odometry on Robots," in *proceedings of the 11th International Conference on Intelligent Autonomous Systems (IAS-11)*, 2008.

[162] M. Warren, D. McKinnon, H. He, A. Glover, M. Shiel and B. Upcroft, "Large scale monocular vision-only mapping from a fixed-wing sUAS," in *Field and Service Robotics*, 2014.

[163] M. Burri, J. Nikolic, P. Gohl, T. Schneider, J. Rehder, S. Omari, M. W. Achtelik and R. Siegwart, "The EuRoC micro aerial vehicle datasets," *The International Journal of Robotics Research,* vol. 35, pp. 1157-1163, 2016.

[164] W. Maddern, G. Pascoe, C. Linegar and P. Newman, "1 year, 1000 km: The Oxford RobotCar dataset," *The International Journal of Robotics Research,* vol. 36, pp. 3-15, 2017.

[165] P. Bergmann, R. Wang and D. Cremers, "Online Photometric Calibration of Auto Exposure Video for Realtime Visual Odometry and SLAM," *IEEE Robotics and Automation Letters,* vol. 3, pp. 627-634, 2018.

[166] J. Engel, V. Usenko and D. Cremers, "A photometrically calibrated benchmark for monocular visual odometry," *arXiv preprint arXiv:1607.02555,* 2016.

[167] D. Caruso, J. Engel and D. Cremers, "Large-scale direct slam for omnidirectional cameras," in *Intelligent Robots and Systems (IROS), 2015 IEEE/RSJ International Conference on*, 2015.

[168] G. Pandey, J. R. McBride and R. M. Eustice, "Ford campus vision and lidar data set," *The International Journal of Robotics Research,* vol. 30, pp. 1543-1552, 2011.

[169] T. Hinzmann, T. Stastny, G. Conte, P. Doherty, P. Rudol, M. Wzorek, E. Galceran, R. Siegwart and I. Gilitschenski, "Collaborative 3D Reconstruction Using Heterogeneous UAVs: System and Experiments," in *International Symposium on Experimental Robotics*, 2016.

[170] A. L. Majdik, C. Till and D. Scaramuzza, "The Zurich urban micro aerial vehicle dataset," *The International Journal of Robotics Research,* vol. 36, pp. 269-273, 2017.

[171] D. Schubert, T. Goll, N. Demmel, V. Usenko, J. St{\"u}ckler and D. Cremers, "The TUM VI Benchmark for Evaluating Visual-Inertial Odometry," *arXiv preprint arXiv:1804.06120,* 2018.

[172] J. Sturm, N. Engelhard, F. Endres, W. Burgard and D. Cremers, "A benchmark for the evaluation of RGB-D SLAM systems," in *Intelligent Robots and Systems (IROS), 2012 IEEE/RSJ International Conference on*, 2012.

[173] M. Fallon, H. Johannsson, M. Kaess and J. J. Leonard, "The MIT stata center dataset," *The International Journal of Robotics Research,* vol. 32, pp. 1695-1699, 2013.

[174] A. Handa, T. Whelan, J. McDonald and A. J. Davison, "A benchmark for RGB-D visual odometry, 3D reconstruction and SLAM," in *Robotics and automation (ICRA), 2014 IEEE international conference on*, 2014.

[175] H. Oleynikova, Z. Taylor, M. Fehr, J. Nieto and R. Siegwart, "Voxblox: Building 3d signed distance fields for planning," *arXiv,* pp. arXiv--1611, 2016.

[176] M. Peris, S. Martull, A. Maki, Y. Ohkawa and K. Fukui, "Towards a simulation driven stereo vision system," in *Pattern Recognition (ICPR), 2012 21st International Conference on*, 2012.

[177] Z. Zhang, H. Rebecq, C. Forster and D. Scaramuzza, "Benefit of large field-of-view cameras for visual odometry," in *Robotics and Automation (ICRA), 2016 IEEE International Conference on*, 2016.